\documentclass{article}
\usepackage{spconf,amsmath,graphicx,hyperref}
\usepackage{xcolor}
\usepackage{amssymb}
\definecolor{gitblue}{RGB}{38, 70, 83}

\title{Can Synthetic Images Serve as Effective and Efficient Class Prototypes?}
\name{Dianxing Shi$^{1}$, Dingjie Fu$^{2 \dagger}$, Yuqiao Liu$^{3}$, Jun Wang$^{4 *}$  }
\address{Beijing Research Institute of Uranium Geology$^{1}$, Huazhong University of Science and Technology$^{2}$,\\
The Hong Kong University of Science and Technology (Guangzhou)$^{3}$, Great Bay University$^{4}$ \\
}

\begin{document}
\ninept
\maketitle
\makeatletter
\renewcommand\@makefnmark{}
\makeatother

\footnotetext{${}^{\dagger}$ Project lead; ${}^{*}$ Corresponding author.}
\begin{abstract}
Vision-Language Models (VLMs) have shown strong performance in zero-shot image classification tasks. However, existing methods, including Contrastive Language-Image Pre-training (CLIP), all rely on annotated text-to-image pairs for aligning visual and textual modalities. This dependency introduces substantial cost and accuracy requirement in preparing high-quality datasets. At the same time, processing data from two modes also requires dual-tower encoders for most models, which also hinders their lightweight. To address these limitations, we introduce a ``Contrastive Language-Image Pre-training via Large-Language-Model-based Generation (LGCLIP)" framework. LGCLIP leverages a Large Language Model (LLM) to generate class-specific prompts that guide a diffusion model in synthesizing reference images. Afterwards these generated images serve as visual prototypes, and the visual features of real images are extracted and compared with the visual features of these prototypes to achieve comparative prediction. By optimizing prompt generation through the LLM and employing only a visual encoder, LGCLIP remains lightweight and efficient. Crucially, our framework requires only class labels as input during whole experimental procedure, eliminating the need for manually annotated image-text pairs and extra pre-processing. Experimental results validate the feasibility and efficiency of LGCLIP, demonstrating great performance in zero-shot classification tasks and establishing a novel paradigm for classification. Codes at \href{https://github.com/DianxingShi/LG-CLIP}{\textcolor{gitblue}{\textit{Project Link}}}.
\end{abstract}
\section{Introduction}
\label{sec:intro} 
\begin{figure}
    \centering
    \includegraphics[width=1.0\linewidth]{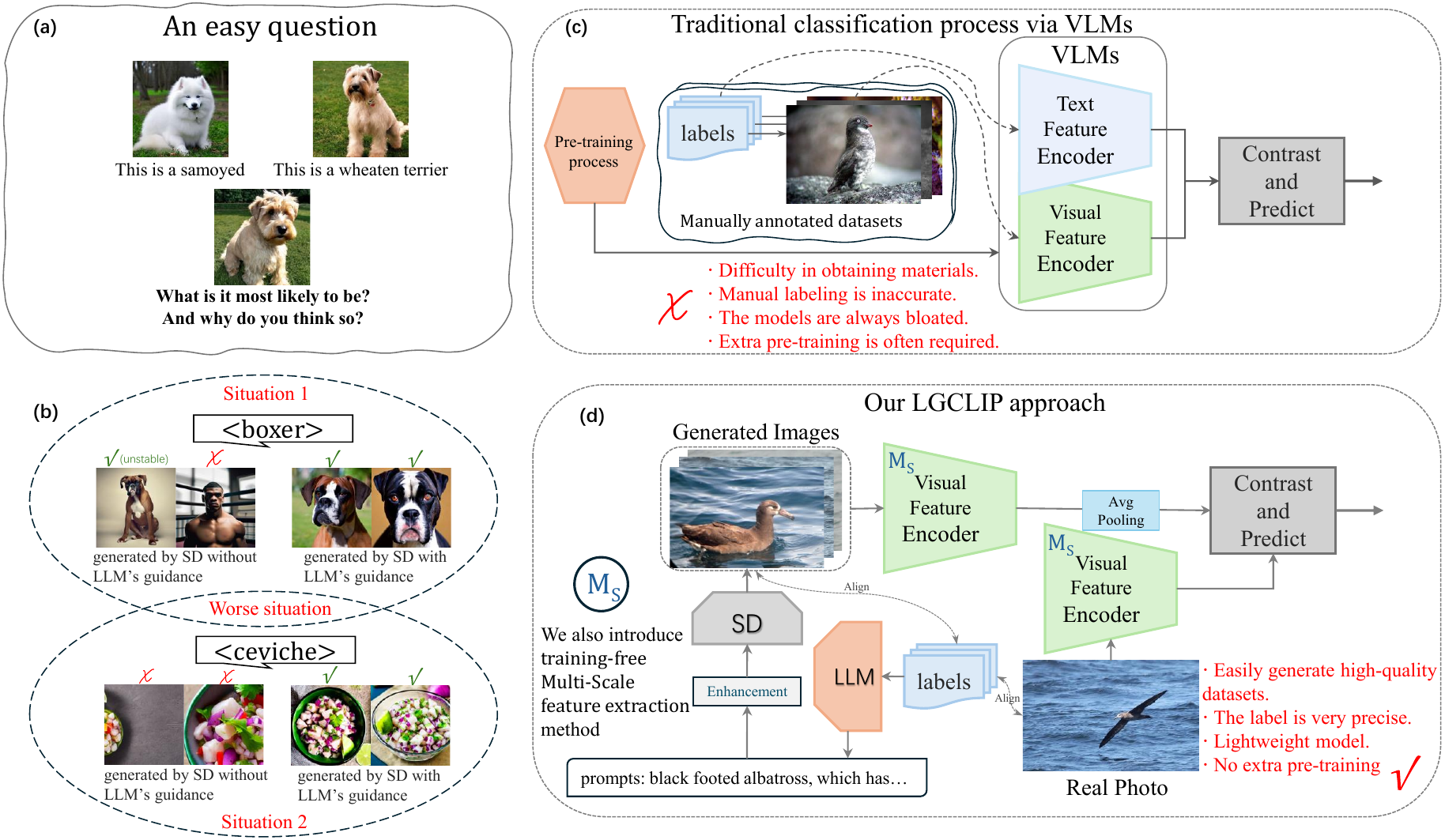}
    \caption{Motivation Illustration. (a) We introduce our motivation with a simple question: ``How do humans classify an object?". (b) Diffusion models are prone to two situations when using text images: i) wrong categories, ii) poor composition. Under the guidance of LLM, this situation can be optimized unsupervised. (c) The process and limitations of image classification methods under the traditional VLM paradigm. (d) The image classification method we proposed and its advantages.}
    \label{fig:1}
\end{figure}
In recent years, deep learning has profoundly transformed computer vision, with the emergence of Vision-Language Models (VLMs) marking a significant paradigm shift \cite{guo2023calip}. Unlike traditional supervised learning methods that rely heavily on labeled data, VLMs learn from large-scale image-text pairs and exhibit unprecedented generalization abilities \cite{udandarao2023sus}. Among them, CLIP \cite{radford2021learning} has achieved remarkable success by aligning image and text features in a shared embedding space, enabling natural language supervision for visual tasks. By simply converting category labels into textual prompts (e.g., ``a photo of a {class}’’), CLIP achieves zero-shot image classification (ZSL) without requiring task-specific training data, thereby reducing dependence on labeled datasets \cite{chen2025interpretable, qian2024online}.

Despite these advances, the core paradigm of image-text alignment presents inherent limitations. Performance depends heavily on the wording of textual prompts, which has led to costly and unstable ``prompt engineering’’ \cite{saha2024improved}. Automated solutions such as CoOp\cite{Zhou_2022} and CoCoOp \cite{Zhou_2022_CVPR} learn soft prompts to adapt across tasks, but they suffer from overfitting, reduced generalization to novel categories, and high computational cost due to optimization through large encoders. Consequently, reliance on text encoders during inference remains a fundamental bottleneck \cite{zhang2022show}.

In parallel, text-to-image (T2I) diffusion models have advanced rapidly, demonstrating the ability to generate highly realistic and diverse images \cite{he2022synthetic,fu2024discriminative}. These models have been applied successfully to augment training datasets with synthetic samples, improving classifier robustness and zero-shot ability \cite{azizi2023syntheticdatadiffusionmodels}. Recent research further combines diffusion models with large language models (LLMs) to overcome the limitations of simple category prompts. LLMs can generate detailed and unbiased descriptions \cite{qu2023layoutllm}, guiding diffusion models to synthesize higher-quality and more controllable images. Such LLM-guided generation has been shown to improve data diversity and reduce semantic bias, thereby enhancing downstream classification performance \cite{qin2024diffusiongpt}.

This raises a fundamental question: rather than attempting to patch the fragile text-link in VLMs, can we leverage the generative power of LLMs and T2I models to completely eliminate the reliance on text encoders during inference, enabling a more robust and efficient purely visual classification paradigm? Or in other words: \textbf{Can synthetic images serve as effective and efficient class prototypes?}

To address this challenge, we introduce Contrastive Language-Image Pre-training via Large-Language-Model-based Generation (LGCLIP), an innovative training-free zero-shot classification framework. LGCLIP departs from traditional image-text alignment and instead adopts a real image–generated image contrast strategy. Specifically, LLMs first produce detailed, unbiased prompts that drive diffusion models to synthesize category-specific reference images, termed ``visual prototypes.’’ During inference, a lightweight visual encoder extracts features from both real images and prototypes, and classification is performed via feature similarity. This design shifts the computational burden of LLMs and diffusion models to an offline, one-time data preparation stage, ensuring high efficiency at inference. Our main contributions are as follows:
\begin{itemize}
\item We propose a novel pure visual feature classification paradigm for zero-shot learning, which avoids the ambiguity and sensitivity of text prompts by comparing real images directly with generated visual prototypes.
\item We exploit the language and knowledge capabilities of LLMs to automatically produce high-quality prompts for diffusion models, enabling unbiased dataset construction and significantly enhancing model generalization.
\item Our comprehensive and rigorous experimental results demonstrate that synthetic images can be used as effective and efficient visual prototypes to help image classification tasks. The experimental results also show promising performance.
\end{itemize}

\section{RELATED WORK}
\label{sec:format}
\textbf{VLM Architectures.} CogVLM\cite{CogVLM} integrates a trainable "visual expert module" at each Transformer layer to solve the problem of insufficient fusion between visual and language features in shallow alignment architectures. LLaVA\cite{LLaVA} uses a simple linear layer to project visual features into the language model's word embedding space, effectively connecting pre-trained vision and language models with minimal training overhead for visual instruction fine-tuning. \\
\textbf{Optimizing Text-Image pairs for images.} CoDA\cite{CoDA} uses contrastive visual data augmentation to guide the diffusion model in generating targeted synthetic images, addressing poor performance in recognizing novel or confused concepts due to data insufficiency . MosaicFusion\cite{mosaicfusion} uses a diffusion model to generate multiple controllable objects and segmentation masks across different regions of a single image canvas, solving the issue of limited labeled data in instance segmentation tasks, especially in long-tail and open-set scenarios . 


\section{Methodology}
\label{sec:method}

\begin{figure}
    \centering
    \includegraphics[width=1.0\linewidth]{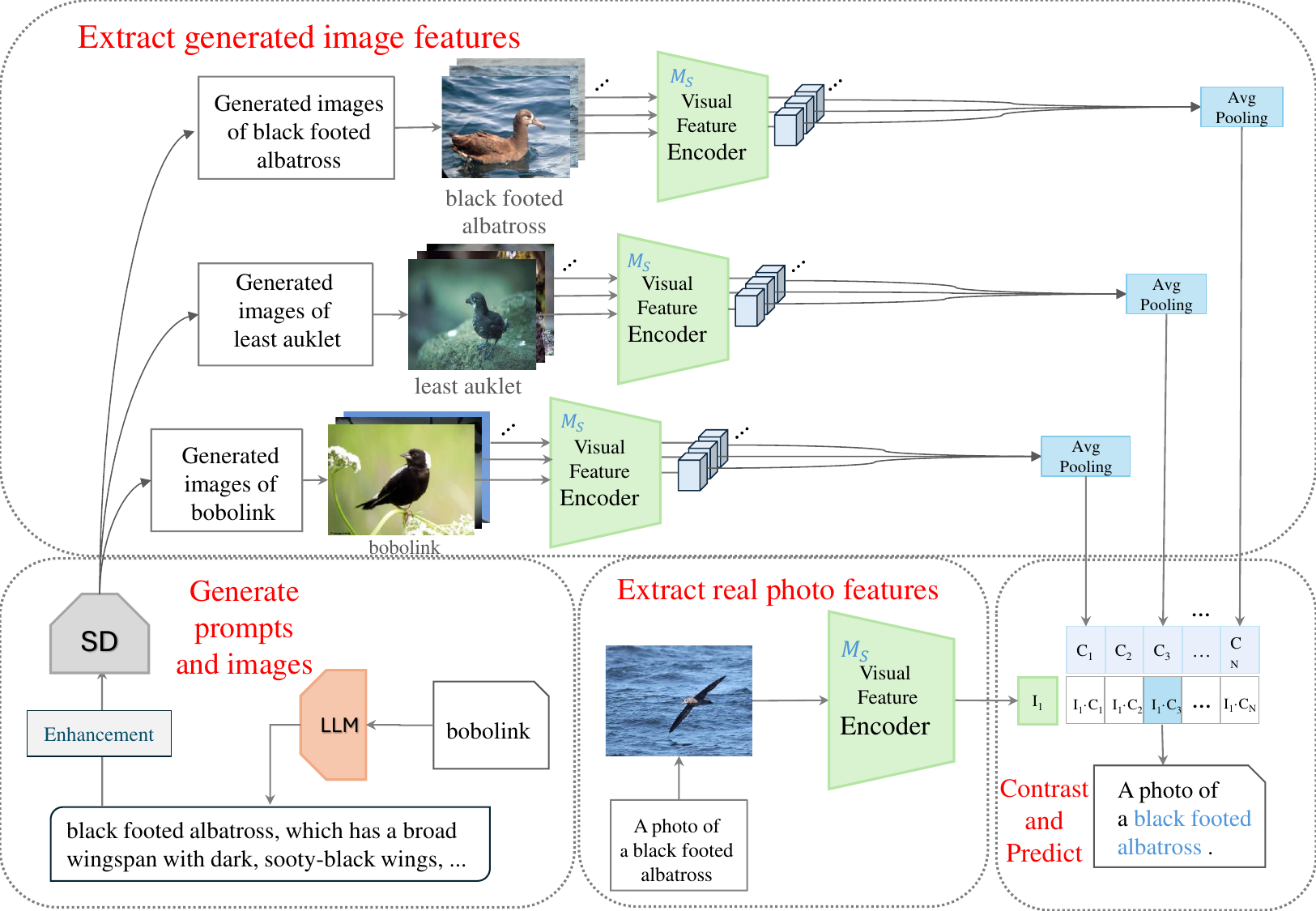}
    \caption{Our proposed LGCLIP workflow.}
    \label{fig:2}
\end{figure}

Our method leverages a Large Language Model (LLM), Stable Diffusion (SD), and a pretrained Visual Feature Encoder (VFE) to perform a complete image classification task. The specific operational process is as follows and in Figure \ref{fig:2}:\\
\textbf{i)} Call the relatively mature and reliable large language model API at this stage to guide it to generate several prompt words (sentences) with different focuses for a specific class.\\
\textbf{ii)}  Input the generated more comprehensive and detailed prompt words into the diffusion model to guide it to generate images.\\
\textbf{iii)}  Use the multi-scale visual feature extraction module to extract the multi-scale visual features of the real images and the generated images of the corresponding classes in the commonly used image classification datasets, and use cosine similarity comparison to complete the classification task of the real images.

\subsection{LLM generates prompts}
The main advantage of the large language model in our method is its ability to retrieve vast information from the Internet and generate relatively accurate prompt words, despite the presence of both reliable and unreliable sources. However, since common image classification tasks often match ``\texttt{a photo of a <broad category>}'' with an image using training-free VLMs like CLIP, it is challenging to identify the key features the model uses for recognition. To generate unbiased prompts, we adopt the ``\textbf{coarse-grained to fine-grained multi-layer splicing}'' method. Specifically, when providing a global task prompt, we instruct the model as follows:
\begin{quote}
\itshape
$<$system$>$\\
``Write a detailed description of \texttt{\{class\_name\}}, including its unique features. The description I need is for image generation, so the description you give must be a clear visual feature that can help the generator understand the content and reduce ambiguity to the greatest extent."
\end{quote}

Furthermore, we tell the model:
\begin{quote}
\itshape
``Try to imitate the process of human eyes recognizing objects to classify the features of objects from coarse to fine granularity and generate prompt words. You need to estimate your own adjectives to ensure that the features generated by your adjectives in the generated image are not distorted or exaggerated. The input length should remain roughly the same, and only generate descriptions of newly updated classes each time to avoid repeating descriptions of classes that have already been generated above." 
\end{quote}
The specific process, including output normalization and unbiased guidance, is a little more complicated, but the basic principle is as above. After obtaining these prompt words, we simply process them and use them to generate images in the diffusion model.
\subsection{Diffusion model generates images}
After obtaining the processed prompts, we employ a diffusion model to generate images. Among various generative approaches—such as GANs, VAEs, and diffusion models—we adopt Stable Diffusion (SD) for its balanced trade-off between controllability, efficiency, and image quality. Consistent with our core motivation of ``using unbiased generated images as the basis for classifying real images,’’ we select SD 2.1 as the generation engine, as it produces realistic-style images with strong text-to-image capability. To ensure reproducibility, a fixed global random seed is applied.
\subsection{ Visual Feature Encoder extracts visual features}
CLIP’s VFE effectively captures deep image features and fine-grained details. We exploit this to extract visual features from both generated and real images, which often contain complex compositions. To maintain a training-free, lightweight framework, we propose a training-free multi-scale feature extraction method:  

Global features are obtained via a single VFE backbone. In our training-free setting, ``multi-scale'' means extracting features from cropped regions, where each scale is defined by a weight $W_n$ and a cropping ratio $CR_n$, yielding one feature vector per region. Aggregating features across scales captures the visual focal points of complex images. Formally, the final representation is:
\begin{equation}
\text{Fs} = \sum_{n=1}^{N} W_n \cdot F_n, \\
\end{equation}
\begin{equation}
W_n = \frac{N+1-n}{\sum_{i=1}^{N} i},  \\    
\end{equation}
\begin{equation}
F_n = VFE(B_k, CR_n, I_p), \quad
CR_n = \frac{1}{n}    
\end{equation}

where $N$ represents the multi-scale resolution level, $I_p$ is the input image and $B_k$ is the VFE backbone. This compact multi-scale representation enhances the extraction of comprehensive features from both generated and real images, which we validate in Section~\ref{sec:Ablation}.

\subsection{Image Classification via Contrastive Prediction}
As described previously, for each class \(c\), we construct a text prompt from its name and description, then generate a synthetic image \(\widetilde{I}_c\) via a diffusion model. The CLIP image encoder \( \mathrm{VFE}(\cdot) \) extracts its feature vector \(\mathrm{Fs}_c \in \mathbb{R}^d\), serving as the class prototype. For \(N_g\) generated images of the same class, we compute the averaged prototype:
\begin{equation}
\mathrm{Fs}_c^{N_g} = \frac{1}{N_g} \sum_{j=1}^{N_g} \mathrm{Fs}_{c,j}
\end{equation}

For each test image \(\widetilde{I}_i\), we similarly obtain its feature \(\mathrm{Fs}_i \in \mathbb{R}^d\). The similarity to each class prototype is measured via inner product:
\begin{equation}
S_{i,c} = \mathrm{Fs}_i^\top \mathrm{Fs}_c^{N_g}, \\
\end{equation}
\begin{equation}
\widehat{y}_i = \arg\max_{c} S_{i,c},  \quad \\    
\end{equation}
\begin{equation}
\mathrm{Acc} = \frac{1}{N} \sum_{i=1}^{N} \mathbb{I}(\widehat{y}_i = y_i)
\end{equation}
where \(\mathbb{I}(\cdot)\) is the indicator function returning 1 if the prediction is correct, and 0 otherwise.

\section{Experiments}
\label{sec:typestyle}
Our experimental process follows the sequence outlined in Section \ref{sec:method}. In this chapter, we provide a detailed description of each stage of the procedure, highlighting key parameters, and present our experimental results followed by an analysis leading to conclusions.
\subsection{Dataset}
To conduct a comprehensive evaluation, we selected six widely-used datasets for zero-shot learning, encompassing a broad range of objects, scenes, and fine-grained categories: CUB \cite{wah2011caltech}, FLO \cite{FLO}, PET \cite{PET}, ImageNet \cite{imagenet}, FOOD \cite{FOOD}, and EUROSAT \cite{eurosat}. These datasets serve as the foundational data for the entire experimental process.Since the training process is not involved, we use all the original images from each dataset as the test set.
\subsection{Experimental Setup}
\textbf{Prompt generation.} Following the previous methodology, we call the Grok 3.0 API using the default temperature and top\_p to generate ten composite prompts per class from original labels and documented any obvious biases. For example, for \texttt{<Boxer>} in the PET dataset, the LLM may produce: 
\begin{quote}
\itshape
“a muscular boxer wearing red gloves, standing in a boxing ring, fists raised, intense expression.”
\end{quote}
This may yield images of a human boxer or occasionally a humanoid boxer dog. No extra context is provided, allowing direct evaluation of the LLM's understanding and self-correction.\\
\textbf{Image generation.} We use Stable Diffusion 2.1, storing prompts as \texttt{[dataset/class\_name/No.]}, with each producing one image. Each class generates $N_g$ images, and only several key parameters (e.g., \texttt{guidance\_scale}, \texttt{num\_inference\_steps}) are adjusted for quality and efficiency.\\
\textbf{Feature extraction and contrastive prediction.} Visual features are extracted using CLIP backbones (RN50 \cite{RN50}, ViT-B/32, ViT-B/16, ViT-L/14 \cite{ViT}). For an image $I_p$, we compute a multi-scale weighted feature $F_s^p$. Features from $N_g$ images are sampled and average-pooled to form the prototype of class $c$. A contrastive score matrix assigns each real image to the class with highest similarity, and final performance is reported using macro accuracy.

\subsection{Results}
\begin{figure}[ht]
    \centering
    \includegraphics[width=0.8\linewidth]{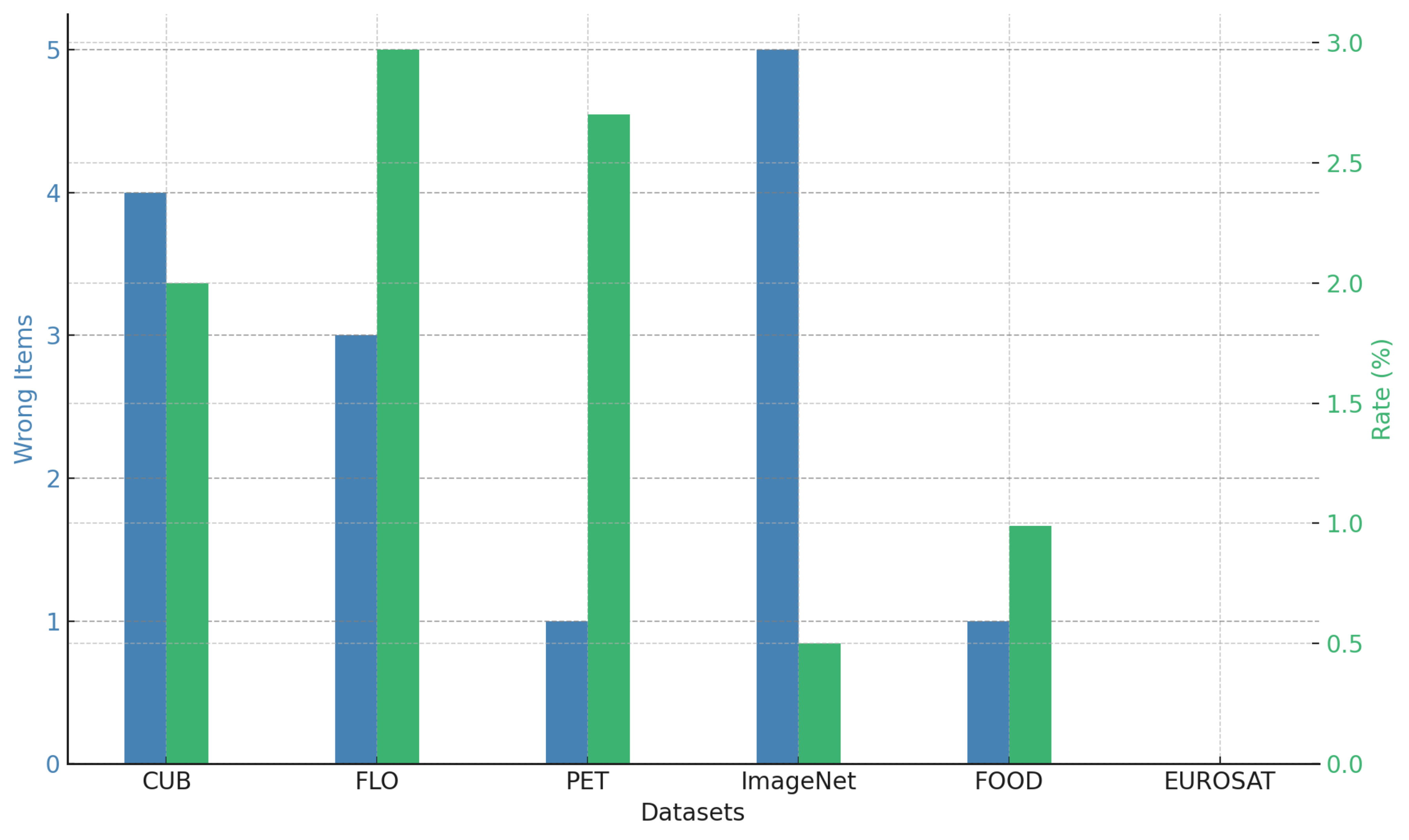}
    \caption{Error statistics: The final error entry and category ratio for each data set. Here, when there is an intersection between prompt errors and images errors, it is only counted once.}
    \label{fig:4}
\end{figure}

Since our approach does not follow the conventional Text-to-Image (T2I) paradigm for classification, its results are not directly comparable to T2I-based methods. Moreover, we do not specifically investigate differences among various visual feature encoders; instead, all experiments are conducted using four CLIP-based backbones. Results of the non-intervention generation experiments are reported in Table \ref{table:1}, while those of the coarse-grained human intervention experiments are shown in Table \ref{table:2}. We also tracked errors from prompt generation to image generation across all six datasets, summarized in Figure \ref{fig:4}. As shown, our method consistently outperforms the baseline across all backbones and datasets. The gain is smallest on ImageNet, while EUROSAT shows the most pronounced improvement, achieving up to a 192\% increase over the baseline.
\begin{table}[ht]
\centering
\caption{Experimental results of our method of accuracy on six datasets using four different backbones. There is \textbf{no correction} here for the wrong prompts generated by LLM or the wrong pictures generated by SD (obviously wrong categories). CLIP’s baseline results are obtained by directly extracting features and comparing predictions between generated images and real photos using CLIP’s \( \mathrm{VFE}(\cdot) \). }
\vspace{0.5em}
\scriptsize
\begin{tabular}{c c c c c c c c c c}
\hline
\textbf{Backbone} & \multicolumn{3}{c}{\textbf{CUB}} & \multicolumn{3}{c}{\textbf{FLO}} \\
\cline{2-7}
 & Ours & CLIP & $\Delta$ & Ours & CLIP & $\Delta$ \\
\textbf{RN50} & 34.55 & 32.90 & \textcolor{red}{+1.65} & 33.58 & 31.87 & \textcolor{red}{+1.71} \\

\textbf{ViT-B/32} & 44.98 & 42.03 & \textcolor{red}{+2.95} & 45.96 & 38.98 & \textcolor{red}{+6.98} \\

\textbf{ViT-B/16} & 49.57 & 45.96 & \textcolor{red}{+3.61} & 50.06 & 43.73 & \textcolor{red}{+6.33} \\

\textbf{ViT-L/14} & 57.47 & 54.07 & \textcolor{red}{+3.40} & 61.35 & 51.93 & \textcolor{red}{+9.42} \\
\hline
\end{tabular}

\begin{tabular}{c c c c c c c c c c}
\hline
\textbf{Backbone} & \multicolumn{3}{c}{\textbf{PET}} & \multicolumn{3}{c}{\textbf{ImageNet}} \\
\cline{2-7}
 & Ours & CLIP & $\Delta$ & Ours & CLIP & $\Delta$ \\
\textbf{RN50} & 57.05 & 50.94 & \textcolor{red}{+6.11} & 39.26 & 37.45 & \textcolor{red}{+1.81} \\
\textbf{ViT-B/32} & 63.75 & 60.26 & \textcolor{red}{+3.49} & 46.72 & 44.29 & \textcolor{red}{+2.43} \\
\textbf{ViT-B/16} & 67.18 & 65.14 & \textcolor{red}{+2.04} & 51.89 & 49.56 & \textcolor{red}{+2.33} \\
\textbf{ViT-L/14} & 77.43 & 72.74 & \textcolor{red}{+4.79} & 59.67 & 56.49 & \textcolor{red}{+3.18} \\
\hline
\end{tabular}

\begin{tabular}{c c c c c c c c c c}
\hline
\textbf{Backbone} & \multicolumn{3}{c}{\textbf{FOOD}} & \multicolumn{3}{c}{\textbf{EUROSAT}} \\
\cline{2-7}
 & Ours & CLIP & $\Delta$ & Ours & CLIP & $\Delta$ \\
\textbf{RN50} & 52.61 & 51.12 & \textcolor{red}{+1.49} & 34.67 & 11.84 & \textcolor{red}{+22.83} \\

\textbf{ViT-B/32} & 63.17 & 60.15 & \textcolor{red}{+3.02} & 32.42 & 22.22 & \textcolor{red}{+10.20} \\

\textbf{ViT-B/16} & 71.13 & 69.97 & \textcolor{red}{+1.16} & 42.43 & 18.31 & \textcolor{red}{+24.12} \\

\textbf{ViT-L/14} & 78.98 & 75.40 & \textcolor{red}{+3.58} & 56.20 & 27.78 & \textcolor{red}{+28.42} \\
\hline
\end{tabular}

\label{table:1}
\end{table}



\begin{table}[ht]
\centering
\caption{The accuracy chart after manual calibration to generate all coarse-grained error categories into correct categories. It is practically meaningful together with Table 1 so we list them separately.}
\vspace{0.5em}
\scriptsize
\begin{tabular}{c c c c c c c}
\hline
\textbf{Backbone} & \multicolumn{3}{c}{\textbf{CUB}} & \multicolumn{3}{c}{\textbf{FLO}} \\
\cline{2-7}
 & Ours & CLIP & $\Delta$ & Ours & CLIP & $\Delta$ \\

\textbf{RN50} & 34.84 & 32.90 & \textcolor{red}{+1.94} & 33.71 & 31.87 & \textcolor{red}{+1.84} \\

\textbf{ViT-B/32} & 44.98 & 42.03 & \textcolor{red}{+2.95} & 45.96 & 38.98 & \textcolor{red}{+6.98} \\

\textbf{ViT-B/16} & 49.70 & 45.96 & \textcolor{red}{+3.74} & 50.45 & 43.73 & \textcolor{red}{+6.72} \\

\textbf{ViT-L/14} & 60.36 & 54.07 & \textcolor{red}{+6.29} & 62.39 & 51.93 & \textcolor{red}{+10.46} \\
\hline
\end{tabular}

\begin{tabular}{c c c c c c c}
\hline
\textbf{Backbone} & \multicolumn{3}{c}{\textbf{PET}} & \multicolumn{3}{c}{\textbf{ImageNet}} \\
\cline{2-7}
 & Ours & CLIP & $\Delta$ & Ours & CLIP & $\Delta$ \\

\textbf{RN50} & 57.61 & 50.94 & \textcolor{red}{+6.67} & 39.37 & 37.45 & \textcolor{red}{+1.92} \\

\textbf{ViT-B/32} & 63.75 & 60.26 & \textcolor{red}{+3.49} & 46.84 & 44.29 & \textcolor{red}{+2.55} \\

\textbf{ViT-B/16} & 65.68 & 65.14 & \textcolor{red}{+0.54} & 51.93 & 49.56 & \textcolor{red}{+2.37} \\

\textbf{ViT-L/14} & 78.04 & 72.74 & \textcolor{red}{+5.30} & 59.74 & 56.49 & \textcolor{red}{+3.25} \\
\hline
\end{tabular}

\begin{tabular}{c c c c c c c}
\hline
\textbf{Backbone} & \multicolumn{3}{c}{\textbf{FOOD}} & \multicolumn{3}{c}{\textbf{EUROSAT}} \\
\cline{2-7}
 & Ours & CLIP & $\Delta$ & Ours & CLIP & $\Delta$ \\

\textbf{RN50} & 53.11 & 51.12 & \textcolor{red}{+1.99} & 34.67 & 11.84 & \textcolor{red}{+22.83} \\

\textbf{ViT-B/32} & 63.43 & 60.15 & \textcolor{red}{+3.28} & 32.42 & 22.22 & \textcolor{red}{+10.20} \\

\textbf{ViT-B/16} & 71.67 & 69.97 & \textcolor{red}{+1.70} & 42.43 & 18.31 & \textcolor{red}{+24.12} \\

\textbf{ViT-L/14} & 79.25 & 75.40 & \textcolor{red}{+3.85} & 56.20 & 27.78 & \textcolor{red}{+28.42} \\
\hline
\end{tabular}

\label{table:2}
\end{table}

\begin{table}[ht]
\centering
\caption{Ablation experiment results. \textbf{CLIP} means experimental results obtained by only using CLIP's VFE loaded with pre-trained weights. $LLM$ indicates the use of LLM to generate prompts, and $M_S$ indicates the use of multi-scale feature extraction method. The best results are indicated in bold.}
\vspace{0.5em}
\scriptsize
\begin{tabular}{c c c c c}
\hline
\textbf{Methods} & \textbf{RN50} & \textbf{ViT-B/32} & \textbf{ViT-B/16} & \textbf{ViT-L/14} \\
\cline{2-5}
\textbf{CLIP} & 37.45 & 44.29 & 49.56 & 56.49 \\
\textbf{CLIP\&$LLM$} & 36.51 & 42.17 & 50.03 & 57.31 \\
\textbf{CLIP\&$M_S$} & 38.75 & 46.32 & \textbf{52.27} & 58.56 \\
\textbf{CLIP\&$LLM$\&$M_S$} & \textbf{39.26} & \textbf{46.72} & 51.89 & \textbf{59.67} \\
\hline
\end{tabular}

\begin{tabular}{c c c c c}
\hline
\textbf{Methods} & \textbf{RN50} & \textbf{ViT-B/32} & \textbf{ViT-B/16} & \textbf{ViT-L/14} \\
\cline{2-5}
\textbf{CLIP} & 11.84 & 22.22 & 18.31 & 27.78 \\
\textbf{CLIP\&$LLM$} & 27.16 & 27.84 & 40.20 & 47.74 \\
\textbf{CLIP\&$M_S$} & 18.74 & 31.91 & 36.02 & 53.04 \\
\textbf{CLIP\&$LLM$\&$M_S$} & \textbf{34.67} & \textbf{32.42} & \textbf{42.43} & \textbf{56.20} \\
\hline
\end{tabular}

\label{table:3}
\end{table}

\subsection{Ablation Study}
\label{sec:Ablation}
We conducted ablation studies along two dimensions using EUROSAT and ImageNet, representing the largest and smallest average performance gains, respectively. Table \ref{table:3} shows the impact of prompt generation and multi-scale feature extraction ($M_S$), while backbone choice also matters: larger pre-trained models with more parameters exhibit stronger VFE capability.

On ImageNet, $M_S$ provides more improvement than LLM, which sometimes degrades performance due to many visually similar categories. In contrast, LLM significantly benefits EUROSAT, where detailed prompts help the SD model generate more accurate images and reduce noisy labels. Additional datasets (not shown) confirm that LLM aids fine-grained tasks by producing more discriminative prompts.

The complexity of natural images in ImageNet makes it hard for an untrained $M_S$ module to capture key features, whereas the structured EUROSAT imagery makes multi-scale extraction particularly effective, explaining the strong gains. Although real-world images remain challenging, our training-free $M_S$ still provides notable improvements.

\subsection{Limitations and expectations}
Based on the experimental results and ablation study, we can summarize those main limitations:\\

There is still a gap in accuracy compared to traditional methods like CLIP's T2I method \cite{radford2021learning}, and the performance varies greatly across different datasets. This is the main deficiency. We look forward to optimizing each aspect of our method in the future to achieve performance comparable to traditional methods.\\

Our multi-scale feature extraction method operates independently of the pre-trained VFE. Adhering to our philosophy of being entirely training-free, the generalization of our method is insufficient, making it undoubtedly challenging to face more complex task scenarios. We aspire to explore a more generalized multi-scale feature extraction method, ideally one that remains training-free.

\section{Conclusion}
\label{sec:majhead}
In this paper, we review the fundamental processes and limitations of traditional image classification. To address these issues, we propose a novel approach that leverages large language models (LLMs) to generate precise prompts, guiding generative models to create high-quality, unbiased datasets. Our method performs training-free, zero-shot image classification using only visual features, avoiding text-image alignment and standing apart from existing methods. We also explore using generative models for prototype creation and LLM-guided prompt generation. Importantly, our model requires no  extra training.\\ 
\textbf{Future Work.} Our method relies on existing base models and has not been validated across a broader set of foundational models. The training-free multi-scale feature extraction also limits generalization. Future work will focus on: i) Evaluating a wider range of base models. ii) Optimizing multi-scale feature extraction and integrating visual feature modules under other VLMs.

\newpage
\bibliographystyle{IEEEbib}
\bibliography{strings,refs}

\end{document}